\documentclass[sigconf,screen]{acmart}
\usepackage{pifont}
\usepackage{CJKutf8}
\AtBeginDocument{%
  \providecommand\BibTeX{{%
    \normalfont B\kern-0.5em{\scshape i\kern-0.25em b}\kern-0.8em\TeX}}}

\setcopyright{acmcopyright}
\copyrightyear{2021}
\acmYear{2021}
\usepackage{comment}

\newcommand{\bo}{\mathbf{o}}

\newcommand{\bx}{\mathbf{x}}

\newcommand{\bW}{\mathbf{W}}

\newcommand{\bb}{\mathbf{b}}

\begin{document}

\title{Multi-Modal Attribute Extraction for E-Commerce}

\author{Aloïs De la Comble, Anuvabh Dutt, Pablo Montalvo, Aghiles Salah}
\authornote{Alphabetical order. All authors contributed equally to this research.}
\email{{alois.delacomble, anuvabh.dutt, pablo.a.montalvo, aghiles.salah}@rakuten.com}
\affiliation{%
 \institution{Rakuten Group, Inc.}
 \city{Paris}
 \country{France}
}

\renewcommand{\shortauthors}{RIT Paris}

\begin{abstract}
To improve users' experience as they navigate the myriad of options offered by online marketplaces, it is essential to have well-organized product catalogs. One key ingredient to that is the availability of product \emph{attributes} such as color or material. However, on some marketplaces such as Rakuten-Ichiba, which we focus on, attribute information is often incomplete or even missing. 
One promising solution to this problem is to rely on deep models pre-trained on large corpora to predict attributes from unstructured data, such as product descriptive texts and images (referred to as modalities in this paper). 
However, we find that achieving satisfactory performance with this approach is not straightforward but rather the result of several refinements, which we discuss in this paper. We provide a detailed description of our approach to attribute extraction, from investigating strong single-modality methods, to building a solid multimodal model combining textual and visual information.
One key component of our multimodal architecture is a novel approach to seamlessly combine modalities, which is inspired by our single-modality investigations. In practice, we notice that this new modality-merging method may suffer from a \emph{modality collapse} issue, i.e., it neglects one modality. Hence, we further propose a mitigation to this problem based on a principled regularization scheme.
Experiments on Rakuten-Ichiba data provide empirical evidence for the benefits of our approach, which has been also successfully deployed to Rakuten-Ichiba. We also report results on publicly available datasets showing that our model is competitive compared to several recent multimodal and unimodal baselines.

 %Online marketplaces have large catalogs composed of millions of products. In consequence, to improve users' experience as they navigate this myriad of options, it is essential to have a well-organized catalog. One key ingredient to that is the availability of product \emph{attributes} such as color or material. However, on some marketplaces such as Rakuten-Ichiba, which we focus on, sellers can be small shops as well as massive retailers gathering products from multiple factories, and the large majority do not include attribute information. Hence it is of practical importance to automatically extract product attributes. Items are usually described by different modalities in the form of textual and visual information. In this paper, we introduce our approach for predicting product attributes based on descriptive texts and images, from investigating strong single-modality methods, to building a multimodal model combining textual and visual information. We introduce a novel modality-merging approach, which, for every product, helps the model decide which modality to attend to. Experiments on Rakuten-Ichiba data provide empirical evidence for the benefits of our approach and demonstrate the effectiveness of our modality merging method. The proposed multimodal model also performs well under human evaluations and is being productionized in the Rakuten-Ichiba marketplace. Moreover, empirical results on two publicly available datasets show that our multimodal model offers competitive performance compared to several recent multimodal and unimodal baselines.
  \end{abstract}



%\ccsdesc[500]{Computing methodologies~Neural networks}
%\ccsdesc[500]{Computing methodologies~Machine learning}
%\ccsdesc[500]{Applied computing~E-commerce infrastructure}


\maketitle

\section{Introduction}\label{sec:intro}
%Online marketplaces do not share the limitations of physical stores, and they can offer millions of products.
Relieved from the inventory limitation of physical storefronts, online marketplaces can offer a massive array of products numbering in millions. 
Structuring this sea of options according to common product \emph{attributes}, such as color, material, brand, etc., is vital because it helps users find the products they need more efficiently. 
Moreover, attribute data can serve as a handle to refine search and recommendation results to meet specific needs (e.g., a fashion product that must be of a particular color and material) \cite{Cardoso2018}. 
Unfortunately, information about product attributes is not always available in practice \cite{mauge12}. For instance, on Rakuten Ichiba, which we consider, the large majority of sellers provide only unstructured textual descriptions of their products. It is thereby of great importance to extract attributes automatically from this unstructured data.
Note that attributes are fine-grained information, contrary to product categories or sub-categories such as fashion, shoes, electronics, to name a few. The latter information is also necessary to organize products. However, unlike attributes, product categories are usually indicated by merchants when adding products to the catalog.

We seek to learn from product descriptive texts and images, referred to as modalities, to predict categorical attributes. We frame this problem as a classification task \cite{bishop06, friedman17}. The input is possibly the text, image, or both, of a product. The output (target) is the product's attributes, such as colors. 
Although it is a standard machine learning task with a rich literature, achieving satisfactory results with existing models in real-world environments is not straightforward. In our case, this requires a systematic investigation into several practical research questions, such as \emph{which model to rely on}, \emph{what modality performs best} and \emph{how we should combine multiple modalities}. 

To investigate the above research questions, we first consider cross-modality comparisons. To this end, we leverage pre-trained deep architectures to predict attributes from text or image data. While these models have demonstrated great performance on many classification tasks, achieving good performance in our case is the result of successive refinements and specific modeling choices, which we discuss in Sections \ref{sec:unimodal} and \ref{sec:finetuning}. 
The results from these comparisons turn out to be insightful as they reveal that the best performing modality is product-dependent. This can be imputed in part to the challenging nature of our data. That is, given a product, information about its attributes (e.g., color) are not always reflected in all its descriptive modalities. Moreover, some images may contain multiple instances of the same product under different attribute values (e.g., a handbag under different colors), which would confuse a visual model trying to predict attributes from such images. 
Not having a modality that is always winning motivates pursuing a multimodal approach. However, we find that naively combining textual and visual information may results in sub-optimal performance. Intuitively, for a given product, a modality that is useless for attribute prediction would be a source of noise in a multimodal approach. Following this intuition and inspired by the results of these cross-modality investigations, we propose a new modality merging strategy, which endows the model with extra flexibility to choose the modality to rely on the most for every product. While this new method offers promising results in several cases, we notice that in practice it may suffer from a \emph{modality collapse} issue, in which it almost discards one modality for all products. Hence, we further propose a mitigation to this problem based on principled regularization scheme.

Aside from performance improvement, the proposed modality-merging method contributes towards having a more interpretable multimodal system. Empirical results on Rakuten Ichiba datasets for color and material prediction showcase the benefit of our approach. Furthermore, the proposed multimodal method has successfully passed human evaluations, and it is currently deployed to Rakuten Ichiba. Our main contributions can be summarized as follows.
%We train and evaluate our models on two Rakuten Ichiba datasets for color and material prediction. We focus on assessing the impact of each modality and the importance of the proposed merging method. Interestingly, not only does the result show that our modality merging approach can boost classification performance, but it also contributes towards having a more interpretable multimodal system. Furthermore, the proposed multimodal method went through human evaluations that it passed successfully, and it is currently deployed in Rakuten Ichiba. We also compare our model to multimodal and unimodal baselines on three publicly available datasets. One is a Rakuten dataset made available as a part of a data challenge\footnote{\url{https://challengedata.ens.fr/challenges/59}}, and two are popular datasets commonly used in multimodal studies, namely MM-IMDB \cite{arevalo2017gated} and UPMC FOOD-101 \cite{wang2015food}. Empirical results on these three datasets show that our multimodal model offers competitive performance. 

\begin{itemize}
\item We discuss several refinements that allow us to successfully leverage pre-trained architectures and build solid single-modality models for the task of product attribute prediction.  
\item We propose a new modality merging method inspired by practical results. For every product, it lets the model assign different weights to each modality. We also introduce a principled regularization scheme to mitigate modality collapse.
\item We conduct extensive experiments on five datasets. Our results provide empirical evidence of the effectiveness of the proposed modality merging approach and its regularized version. Moreover, our method offers competitive performance on public datasets compared to several strong unimodal and multimodal baselines.

%Not only it does offer classification performance improvements, but it also yields a more interpretable multimodal model.
%and highest classification performances on the Rakuten Ichiba datasets.
%\item We systematically compare regular concatenation, modality-attention and regularized modality-attention on all datasets present and find the latter method to perform better. We support our claim that modality-attention is a solid approach by showing the attention weights distribution. 
\end{itemize}

% Our main contributions can be summarised as:
% \begin{itemize}
%     \item We train and evaluate our models on two Rakuten Ichiba datasets for color and material prediction. We focus on assessing the impact of each modality and the importance of the proposed merging method. Interestingly, not only does the result show that our modality merging approach can boost classification performance, but it also contributes towards having a more interpretable multimodal system.
%     \item The proposed multimodal method went through human evaluations that it passed successfully, and it is currently being productionized in Rakuten Ichiba.
%     \item We also compare our model to multimodal and unimodal baselines on three publicly available datasets, where empirical results on these three datasets show that our multimodal model offers competitive performance. One is a Rakuten dataset made available as a part of a data challenge\footnote{\url{https://challengedata.ens.fr/challenges/59}}, and two are popular datasets commonly used in multimodal studies, namely MM-IMDB \cite{arevalo2017gated} and UPMC FOOD-101 \cite{wang2015food}. 
% \end{itemize}
We hope our investigation into multimodal attribute extraction, for e-commerce products, will benefit other researchers and practitioners embarking on similar journeys. We will release our code upon paper publication.

\section{Related work}
\label{sec:attribute-extraction}
%Multimodality is an increasingly growing research topic in machine learning with a broad spectrum of models and applications \cite{Baltru2019MultimodalReview}.
In this section, we provide a brief overview of existing multimodal classification methods that are closely related to ours. We focus on the text and image modalities. We do not restrict our discussion to e-commerce data and product attributes extraction. For a detailed review on multimodality, interested readers can refer to \cite{Baltru2019MultimodalReview}.  

Existing methods differ mainly in the base architecture they use to represent each modality and in the way they integrate multiple modalities into a unified approach. Here we discuss existing work along the latter line also referred to as modality fusion or merging in the literature. It is common to distinguish between two major families of approaches, namely \textit{early fusion}, and \textit{late fusion} \cite{Gadzicki2020}. 

\paragraph{Early fusion} Methods in this family consist in processing the data modalities as a whole in such a way as to build a multimodal representation without relying on complex single-modality models. For instance, \citet{kiela2018efficient} convert continuous visual ResNet features into a discrete sequence of tokens that are fed to FastText \cite{joulin16} along with the text tokens. They compared two ways of discretizing the resulting embeddings for memory purposes. 
In a follow-up work \cite{kiela2019}, images are first transformed into ``tokens'' by extracting features from a pre-trained CNN. These tokens are then appended to the text tokens, and this combined input is used to train a BERT model. This method set the previous state-of-the-art on MM-IMDB dataset.
Some works build on the assumption that intermediate features, from either modalitiy, are more important. \citet{Vielzeuf2018} propose a model that has multiple modality merging steps at different layers of the modality encoders. Modality-wise representations are merged into a separate central network that has a joint representation in each layer. Subsequently, \citet{perezrua2019} propose MFAS (Multimodal Fusion Architecture Search), which uses a neural architecture search to determine which fusion layers to use in a given task. \citet{yin2021} further claim that allowing intra-modality feature fusion is a big booster that was not explored in MFAS. Next we discuss methods from the second category, late fusion, to which our contribution belongs to. 
%\vspace{-0.1in}

\paragraph{Late fusion} The methods of this family combine either the class probabilities or the resulting features from unimodal models, into one unified model. Common ways to merge various sources of information in this context are weighted sum or average, and concatenation. For instance, \citet{wang2015food} use TF-IDF vector representations for the text modality and a 19-layer VGG CNN for the image modality. Then the text and visual models are combined by taking a weighted sum of their respective prediction scores, where the weights are determined by cross-validation. \citet{reiter2020} encode an arbitrary number of modalities into vectors using deep neural networks, which are then pooled into a fixed size feature vectors. Note that the best result obtained by this approach also uses a bounding boxes detector in addition to a standard image features extractor.
\citet{jin2021} introduce a way to do efficient model distillation under a multimodal setting. In more details, an auxiliary loss is used for each modality, thereby transferring modality specific information from the teacher to the student. The authors use VisualBERT and TinyBERT, which are both multimodal models. %They report performances of the student network on MM-IMDB and note that building a modality-importance function in the loss was crucial to the student's training. 
\citet{armitage2020} introduce an additional objective for multimodal fusion coming from variational inference. They study the effects of this additional objective and that of different regularization strategies on the MM-IMDB classification task. \citet{arevalo2017gated} use a Gated Multimodal Unit on top of feature extractors that normalize modalities using the {\tt Tanh} activation and then take a weighted sum of the different modality features. %The concatenation operator has to be learned, and the classification module is a 2-layer MLP with maxout activation.
\citet{chen2021multimodalitem} use ViT and BERT, both transformers-based architectures to encode textual and visual information. They combine the text self-attention with a gated text-image cross-attention. They perform experiments on a subset (about 500k products) of the Rakuten Ichiba catalog datasets. Their model predicts high-level product categories, as opposed to ours that focuses on finer-grained attributes. %The authors do not include public datasets for comparison.
It is noted in \cite{singh2020} that pretraining of encoders affects vastly the end performance of multimodal models. They study visio-linguistic pretraining and the importance of the pretraining dataset depending on the downstream task. Best results for MM-IMDB are however achieved without visio-linguistic pretraining as the task is far from other pretraining datasets tasks. \citet{Zahavy2018} considers fusion at the class probability level. The authors introduce a policy that learns to select either the text model of the image one to make the final classification decision.
Similar to our contribution, \citet{zhu20} leverage texts and images to tackle the problem of attribute extraction. To combine the textual and visual features from BERT and ResNet respectively, the authors propose a \emph{cross-modality} attention mechanism, i.e., vanilla attention applied over the textual and visual features (dimensions) simultaneously. This method is different from ours. The aim of the cross-modality attention method is to use the visual information to enhance the text -- sentence -- representation (please see Figure. 2 in \citet{zhu20}), while the goal of our modality-attention is to let the model choose which modality (text or image) to rely on the most for every product.

When taking into account both simplicity of implementation and performance of classification, late fusion techniques are more appealing. Moreover, as state-of-the-art architectures to represent different modalities are constantly evolving, late fusion based methods can be seamlessly updated with the latest models to encode each data modality. In our experiments we compare to several of the above contributions on public benchmark datasets to showcase the benefit of our approach.  
\section{Single Modality Models}
\label{sec:unimodal}
We start our multimodal attribute extraction journey by investigating each modality separately. Without loss of generality, we focus on two modalities: text and image. In this section, we describe our single-modality models and discuss various modeling choices and refinements that allowed us to build solid models of each modality.

\subsection{Text-only model}
\label{sec:text_only}
%For every product, we have access to its title and description. Our goal is to derive text-based representations of the products that are good at predicting their attributes (e.g., color, material). To this end,
We consider a two-level architecture. The first level consists of an adequate model to represent text, while the second one is a multilabel classifier whose aim is to guide the text representation component towards extracting latent features that are good for product attribute prediction.   

\paragraph{Text Representation} We have explored three main models to represent text, namely Bag-of-Word (BoW), Text-CNN \cite{kim-2014-convolutional}, and BERT \cite{Devlin2018}.
In our early experiments, BERT was performing surprisingly worse compared to the former two. However, after several investigations, we found that the bad performance of BERT was caused by the classifier and the fine-tuning process, which has to be done carefully. In the following, we focus on BERT and discuss the refinements that allowed us to obtain the highest results with this model.

BERT, which stands for Bidirectional Encoder Representations from Transformers, is a self-supervised language representation model based on Transformers \cite{Vaswani2017}. It borrows the encoding component of Transformers, i.e., a stack of encoders, each of which is composed of a self-attention layer and a feed-forward network. For more details on the Transformers' architecture, please refer to the original paper \cite{Vaswani2017}. Given a sequence of $N$ tokens $t_1, \ldots, t_N$ as input, BERT produces a list of contextual vector representations $\bx_1^{txt},\ldots,\bx_N^{txt}$, one for each input token, where $\bx_n^{txt}\in\mathbb{R}^{H}$. In our case the input takes the following form:
$$\texttt{[CLS] \{title\} [SEP] \{description\}},$$ 
where \texttt{[CLS]} is the classification token added at the beginning of every input, and \texttt{[SEP]} is a special separator token. The reasons for adding these special tokens will become apparent shortly. 

With the data and model architecture in place, BERT is often pre-trained on unlabeled data using two tasks: masked token imputation and next sentence prediction. It is common practice to rely on pre-trained BERT models, which are readily available on HuggingFace \cite{wolf-etal-2020-transformers}. Hence, from now on, we assume that we have access to a pre-trained BERT. Next, we discuss the classification model, which is key to leveraging BERT to tackle product attribute prediction.       
%Since our texts are mainly in Japanese, we rely on a BERT pre-trained on the Japanese Wikipedia corpus by Tohoku university \cite{cltohoku2021bert}. WeThe pre-training includes masked language modeling and next sentence prediction. Tokenization takes into account latin characters as well as the three scripts in Japanese language, namely kanji, hiragana, and katakana.
%We use the implementation from HuggingFace \cite{wolf-etal-2020-transformers}.

%The data we encode is the concatenation of the title and description of a product. Given the title and description of a product, we create a new string in this manner: \texttt{[CLS] \{title\} [SEP] \{description\}}. This string is given as input to the BERT encoder and it outputs an embedding $x_{text} \in \mathbb{R}^{L \times H}$, with the hidden size $H=768$ as in BERT-base and the sequence length $L$ is at most $512$ non-empty tokens. Shorter inputs are right-padded with zeros. 
\paragraph{Classifier} To leverage BERT for our task, we need to add a target model (a multi-label classifier) on top of it. Two elements are crucial for the choice of the classifier: its input and architecture. 

Regarding the input, a natural choice is to use the representation of the [CLS] token. Indeed, every input sentence/text to BERT serves as a context to represent this token. Therefore the vectors of the [CLS] token can be thought of as the representative of the input texts. Several empirical investigations showed that feeding the [CLS] token for the target model works quite well for some tasks such as sentiment classification \cite{Devlin2018}. There is, however, no evidence that the [CLS] representation will encode useful information for attribute prediction. Hence, we further investigate other possibilities, namely representing a text using the sum or the mean of all its token-vectors.  
In all cases, we obtain a single vector representation for every text, which we denote as $\bx^{txt}$ $\in \mathbb{R}^{H}$.

Whereas some former works have reported successful combinations of BERT with relatively complex models such as LSTMs \cite{gallo20}, in our case, more complex architectures for the classifier led to a substantial performance drop. We find that even passing the BERT-based text representation $\bx^{txt}$ through simple non-linearities, such as RELU and Tanh, causes a significant decrease in classification results. We will discuss this aspect further when we introduce our finetuning procedure in Section~\ref{sec:finetuning}. To overcome this difficulty, we adopt a linear classifier without any hidden layer, 
\begin{eqnarray}
    \label{eq:classifer}
    \tilde{\bo} &=& {\tt Linear}(\bx^{txt}) = \bW\bx^{txt} + \bb \nonumber\\
    \bo&=&
    \begin{cases}
      {\tt Sigmoid}(\tilde{\bo}), & \text{if}\ \text{multilabel objective} \\
      {\tt Softmax}(\tilde{\bo}), & \text{if multiclass objective}
    \end{cases}
\end{eqnarray}
where $\bW$, and $\bb$ denote the weights and bias parameters. This very simple classifier is the best performing one compared to more complex alternatives we have investigated (e.g., deep neural nets).     

%The final embedding is obtained in one of the following two ways:
%\begin{enumerate}
%    \item Summing $x_{text}$ along the sequence dimension, which translates to summing all the \texttt{L} tokens.
%    \item Taking only the \texttt{[CLS]} token representation, which is the first element in $x_{text}$.
%\end{enumerate}
%In both cases, we obtain an embedding vector with the same number of dimensions. We then apply the layer normalisation operation \cite{ba2016layer} obtain a text embedding $x_{text}$ $\in \mathbb{R}^{D_{text}}$.

\subsection{Image-only model}
\label{sec:image_only}
%Similar to the text model, the image-based model we consider consists of an encoder and a classifier.

\paragraph{Image representation} We rely on pre-trained convolutional neural networks, which are standard to learn from images. We consider several variants, including DenseNet \cite{Huang2016} and various ResNet architectures \cite{DBLP:journals/corr/HeZRS15}, and we find DenseNet to give the best results while keeping an affordable computational cost. 
%For encoding image data, we use the Convolutional Neural Network (CNN) architecture DenseNet-121 (121 layers) \cite{Huang2016}. We have tried multiple different CNN architectures such as ResNet-50 and ResNet-152 \cite{DBLP:journals/corr/HeZRS15} and DenseNet-121 gave the best results while keeping an affordable computing cost. 
DenseNet is an architecture inspired by ResNet \cite{DBLP:journals/corr/HeZRS15}. ResNet uses skip connections across blocks of convolutions, which change the function approximated by a block from $F(\bx)$ to $F(\bx) + \bx$, i.e., adding the input to the output of the block. 
%The function approximated is then a perturbation of the identity instead of being close to zero.
The underlying hypothesis is that the function to be approximated by a block is closer to the identity than to the null function. This also makes it easier to backpropagate the gradient as the identity path is unaltered, and thus, it lets the gradient flow to earlier blocks without diminishing its magnitude.
DenseNets builds on ResNet and further implements skip connections across more than one convolutions block. Thus, features from the first blocks can be reused later in the network if needed.

Given a raw image as input, i.e., a three dimensional tensor where dimensions correspond the image channels, width, and height, DenseNet allows us to obtain a continuous vector representation of the image $\bx^{img} \in \mathbb{R}^{D}$.

\paragraph{Classifier} In contrast to BERT, we find that combining DenseNet with relatively complex models, such as deep neural nets, for attribute prediction, does not result in poor performance. However, we do not observe substantial improvements compared to using a shallow network. We thereby adopt a linear model similar to (\ref{eq:classifer}).

\section{Multimodal model}
\label{sec:multimodal}
\begin{figure*}[t]
  \centering
  \includegraphics[scale=0.55]{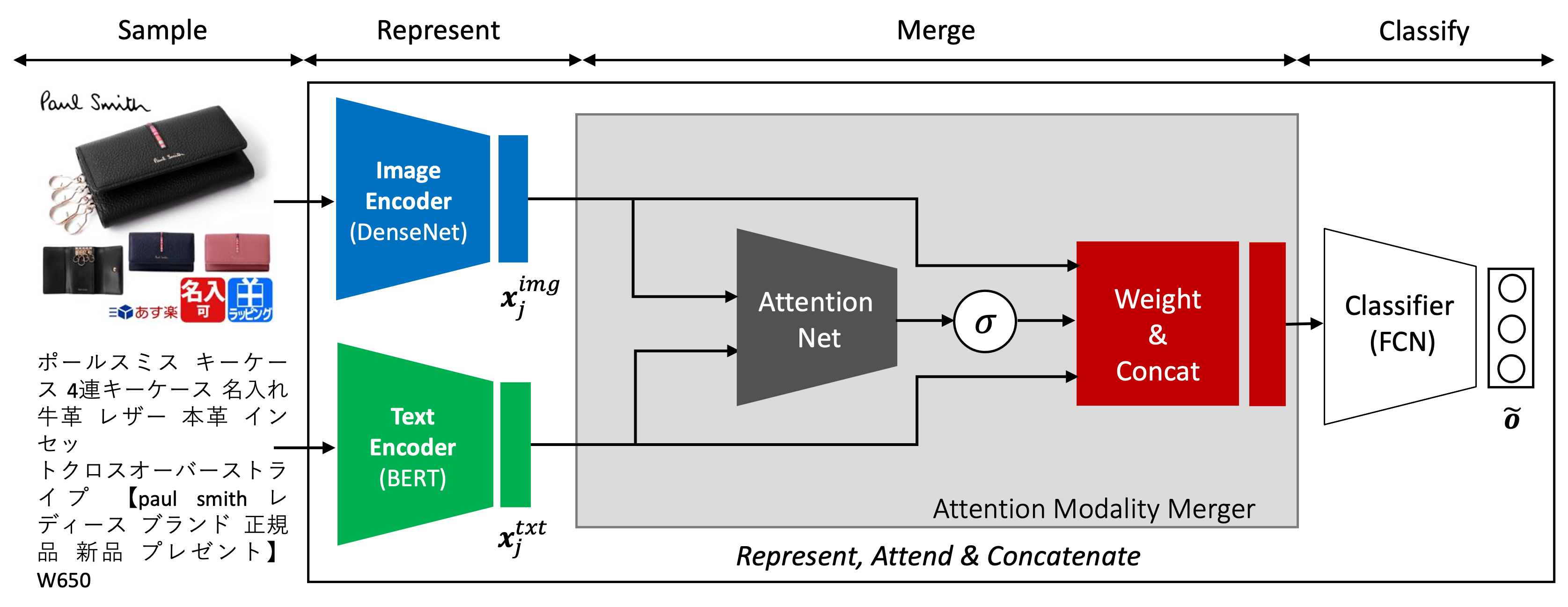}
  \caption[Overview of the proposed multimodal attribute extraction model]{\small Overview of the proposed multimodal attribute extraction model\footnotemark}
  \label{fig:max-system}
\end{figure*}

Cross-modality comparisons show that the text-only model outperforms the image-only model in all cases, if we look at the global results, i.e., the results aggregated on the whole test set, see Tables~\ref{tab:modality-comparison}. However, looking at detailed results, per-product, the text model is not always superior as depicted in Figure \ref{fig:best_modality}. Not having a modality that performs the best on all samples is a strong signal for the importance of pursuing a multimodal approach. In this section, we describe our multimodal model, whose main components are depicted in Figure~\ref{fig:max-system}. We focus on how to combine modalities, which is a long-standing research question in the context of multimodality. 
%
%\begin{figure}[!t]
%  \centering
%  \includegraphics[scale=0.45]{figures/best_performing_modality_color}
%  \caption{\small The number of samples (products) on which each modality performs the best, on the Rakuten Color dataset. The Text/Image legend means both modalities perform equally. Clearly no modality performs the best on all samples.}
%  \label{fig:best_modality_color}
%\end{figure}
%
%
%\begin{minipage}{\linewidth}\centering
\footnotetext{Image and text were gathered from \url{https://item.rakuten.co.jp/rush-mall/paul-psq043/} on 21-09-2021 and resized. Please note that content of the url might vary over time}
%
%\end{minipage}
\subsection{Modality-Attention Merger}
\paragraph{Modality representation} 
Driven by our unimodality investigations, we adopt the text and image encoders described in sections \ref{sec:text_only} and \ref{sec:image_only} to represent the descriptive modalities of every product. %Given the text and image as input, the modality representation module outputs a continuous vector embedding for each modality, $x^{txt}$ and $x^{img}$ respectively.

\begin{figure}
\begin{tabular}{@{}lp{0.4\linewidth}cc@{}}
 \toprule
\small \bf Image & \small \bf Text & \multicolumn{2}{c}{\small \bf Attention} \\
 \midrule
 & & \small Text & \small Image \\
 \midrule
 \raisebox{-\totalheight}{
 \includegraphics[width=0.1\textwidth]{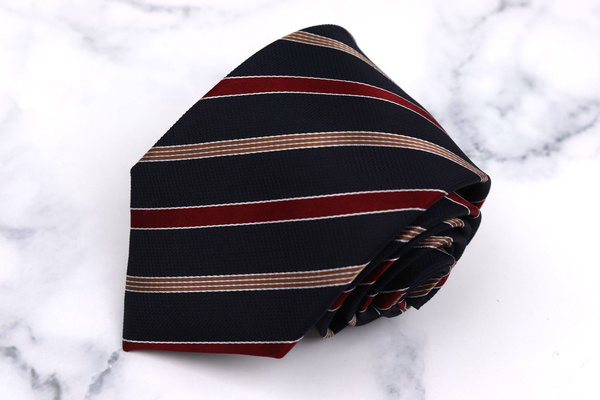}
 } & 
 %\begin{CJK}{UTF8}{min}ボルサリーノ borsalino 日本 シルク ストライプ柄 グレー シルク ブランド ネクタイ 無料 美品 \end{CJK} & 85\% & 15\% \\
 \begin{CJK}{UTF8}{min}\small ボルサリーノ Borsalino ブランド ネクタイ ゆうパケット 送料無料ボルサリーノ Borsalino シルク ストライプ柄 ネイビー シルク ブランド ネクタイ 送料無料 【中古】【美品】品 \end{CJK} & 85\% & 15\% \\

 \midrule
 \raisebox{-\totalheight}{
 \includegraphics[width=0.07\textwidth]{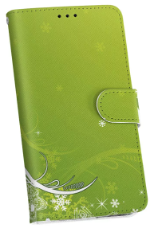}
 } & 
 \begin{CJK}{UTF8}{min}\small shv aquos r アクオス r au エーユー 手帳型 スマホ カバー レザー ケース 手帳タイプ フリップ ダイアリー 二つ折り 革 ラグジュアリー 花 雪の結晶\end{CJK} &
 7\% & 93\%
 \\
\bottomrule
\end{tabular}
\caption[]{
Attention weights computed by the model. The top sample has high text attention, and we see that it has a color 
%("gray", \begin{CJK}{UTF8}{min}グレー\end{CJK})
("Navy", \begin{CJK}{UTF8}{min} ネイビー\end{CJK})
explicitly mentioned in the text. The bottom sample has high image attention (1 - text attention), and there is little color information in the text\footnotemark.}
% two examples of high text attention and high image attention (100 - text attention). The sample with high text attention has a color ("gray") written in the text while the sample with low text attention does not have color information in the text.}
\label{fig:qualitative_example_attention}
\end{figure}

\paragraph{Modality merging}
The cross-modality comparison results, showing that the best-performing modality may be product-dependent, have inspired us to propose a new merging approach, which gives the model the flexibility to decide, for every product, on which modality to focus the most. Formally, given the text and image representations of product $j$, $\bx_j^{txt}$ and $\bx_j^{img}$, we use a shallow neural net to estimate the ``importance'' of each modality:
\begin{eqnarray}
    \label{eq:attention}
    \tilde{p}_j^{txt} &=& {\tt Linear}(\bx_j^{txt}, \bx_j^{img}) = \bW[\bx_j^{txt}, \bx_j^{img}] + \bb \nonumber\\
    p_j^{txt}&=& {\tt Sigmoid}(\tilde{p}_j^{txt})\nonumber\\
    p_j^{img}&=& 1 - p_j^{txt} 
\end{eqnarray}
where $\bW$ and $\bb$ are learnable weight and bias parameters. The modality importance-weights, $p_j^{txt}$ and $p_j^{img}$, are then used to scale the text and image representation before feeding them into the classifier. The input to the classifier takes the following form:
\begin{eqnarray}
    \label{eq:cls_multimodal_input}
    [p_j^{txt} * \bx_j^{txt}, p_j^{img} * \bx_j^{img}],  
\end{eqnarray}
where $[\cdot,\cdot]$ denotes the concatenation operation. Note that the generalization of this approach to more than two modalities is straightforward. We just need to replace the {\tt Sigmoid} output with a {\tt Softmax} one. This approach is similar to the attention mechanism, which is widely used in computer vision and natural language processing models. The difference is that in our case the attention weights are not over tokens or feature-dimensions, they are over modalities. We therefore refer to this method as \emph{\textbf{modality-attention merger}}. Figure~\ref{fig:qualitative_example_attention} shows examples of attention weights learned by the proposed model on Rakuten Ichiba data.

\paragraph{Feature Normalization} We observe that the scale of the vector representations of texts and images can vary substantially. Moreover, for the text model, when using the sum of all token embeddings to represent the whole text, the magnitude of the resulting vector representation can vary depending on the input sequence length. Therefore, the modality whose representation exhibit a higher magnitude would have more impact on the optimization procedure. We can tackle this scale imbalance issue by normalizing the vector representations of the different modalities before feeding them to the attention modality-attention merger. In this work, we rely on LayerNorm \cite{ba2016layer}
%\footnote{Note that one may be tempted to consider Batch Normalization (BatchNorm). The latter strategy, however, normalizes inputs across the batch dimension, which we want to avoid since in our case different samples in a batch can have varying contributions from each of modality.}.
LayerNorm has the benefit of enabling the network to learn the normalization parameters from data, and thus removing the need to search for scaling parameters. 
%Note that one may be tempted to consider Batch Normalization (BatchNorm).
%The latter strategy, however, normalizes inputs across the batch dimension, which we want to avoid since in our case different samples in a batch can have varying contributions from each of modality.

\paragraph{Multimodal classifier} For the same reasons as the ones discussed in section \ref{sec:text_only}, we adopt a shallow classifier similar to the one of equation (\ref{eq:classifer}). The difference is in the input, now it is given by (\ref{eq:cls_multimodal_input}). %To learn or finetune the different components of our modtimodal model jointly, including the classifier, the modality-attention merger, the image and text encoders, we also leverage the same procedure as for training the text-only model, which we introduced in section \ref{sec:text_only}. 

\subsection{Mitigating Modality Collapse}
In practice we find that using this merging strategy offers promising results in several cases. However, we noticed that sometimes (e.g., on Rakuten-Material data in our experiments) this method suffers from a \emph{modality collapse} problem, in which one modality receives a very high weight (close to 1), for almost all products, causing the model to neglect or even discard the other modality. 

To mitigate the modality collapse issue, we propose a principled regularization scheme encouraging the modality-attention distributions to spread their mass across modalities. 
%We achieve this desiderata in a principled way by regularizing the cross-entropy loss of our model using the entropy of the modality attention weights, which is defined as follows:
%\begin{equation}
%    E_{reg} = - p^{txt} \log(p^{txt}) - p^{img}\log(p^{img})
%\end{equation}
Let $p_j = (p_j^{txt}, p_j^{img})$ denote the modality attention distribution for product $j$, and let $q=(\frac{1}{2}, \frac{1}{2})$ be a discrete uniform distribution. We propose to regularize the cross entropy (CE) loss function $\mathcal{L_{\text{CE}}}$ of our model using the Kullback-Leibler (KL) divergence between $p$ and $q$. Our regularized loss function takes the following form:
\begin{equation}
\label{reg_loss}
    \mathcal{L}_{\text{Reg}} = \mathcal{L}_{\text{CE}} + \lambda\times\sum_j\text{KL}(p_j||q),
\end{equation}
where $\lambda$ is a hyperparameter controlling the weight of the above regularization. If $\lambda$ is too high, the modality-attention distribution $p_j$ will match the uniform distribution $q$ giving equal importance to all modalities. In practice we set the value of $\lambda$ based on a validation set. In particular, we are interested in a value of $\lambda$ that allows the modality-attention distribution to deviate from the uniform distribution without falling into the modality collapse scenario. To gain even more insights on why the above regularization scheme would push the model away from skewed attention distributions, it is worth noting the relationship between the above KL terms and the entropy of $p_j$. That is,
\begin{eqnarray}
    \text{KL}(p_j||q) &=& \mathbb{E}_{p_j} [\log{p_j} - \log{q}]\nonumber\\
    &=& p_j^{txt} \log(p_j^{txt}) + p_j^{img}\log(p_j^{img}) + \log{2}\nonumber\\
    &=& - \mathbb{H}(p_j) + \log{2}
\end{eqnarray}
where $\mathbb{H}(p_j)$ is the entropy of the attention distribution for product $j$. Hence minimizing (\ref{reg_loss}) w.r.t. the KL terms is equivalent to maximizing the entropy of the attention distributions, which would encourage the merger towards producing distributions $p_j$ that do not put all their mass on one modality. %We refer to this approach as \emph{ regularized modality-attention merger}. 

\footnotetext{\small Image and text taken from \url{https://item.rakuten.co.jp/daitokka/ss51707/} (top) \url{https://item.rakuten.co.jp/case-style/shv39-001332-nb/} (bottom) on 09-02-2022 and resized. Please note that content of the url might vary over time.}

\section{Training and Finetuning}
\label{sec:finetuning}
Our models comprise pre-trained parameters from DenseNet and BERT, and randomly initialized parameters from the classifiers and merger in the multimodal case. The initial approach we consider is to freeze the pre-trained parameters and train only the randomly initialized parameters. The results of this strategy are however not convincing, which is due to the fact that BERT and DenseNet are pre-trained on tasks different from ours. We therefore choose to jointly fine-tune BERT/DenseNet and learn the parameters of the classifier. For the pre-trained components, we found BERT to be very fragile and requires a gentle fine-tuning, especially in the first iterations.    
%Our architecture comprises both pre-trained parameters from BERT, and randomly initialized parameters from the classifier. The initial approach we consider is to freeze the pre-trained parameters and train the classifier only. The results of this strategy are not convincing, which is due to the fact that BERT is pre-trained on tasks substantially different from ours. We, therefore, choose to fine-tune BERT.
%Our first attempts to do so reveal that the fine-tuning of BERT should be gentle, especially in the first iterations.
In fact, in the beginning, the parameters of the target model (the classifier and merger in the multimodal case) are random, corresponding to inaccurate predictions (high classification error). Thus confident updates of BERT's parameters in early iterations, due to high error backpropagation, cause the model to forget the patterns learned thanks to the pre-training, which results in poor classification performance. We argue that this can also explain why our early investigations with more complex classifiers were unsuccessful, i.e., the harder the target model to train, the less likely we preserve the pre-trained BERT structure. Using relatively small learning rates (e.g., of order 2e-5 using the ADAM update rule) alleviates but does not solve this issue. The most promising approach we investigate in this work is to use a learning rate scheduler, which consists of a warmup period during which the learning rate increases to its maximum, and a cool-down phase in which the rate decreases until the end of the optimization procedure.
%\paragraph{Training and Finetuning} Similar to the text model, we find that training the classifier and updating the encoder parameters jointly offers better results than freezing the pre-trained parameters. This makes sense since the pre-training and target tasks are different. We also find that the training of the image-only model to be less challenging compared to the text-based model. In particular, DenseNet turns out to be quite robust to finetuning as opposed to BERT.

\section{Experiments}
In this section we report the results of our investigation on Rakuten datasets, as well as compare the proposed model to recent multimodal/unimodal baselines on public benchmark datasets.
%We evaluate our model on several real-world dataset. Our goals are two-folds: (i) assess the impact of the different components of our model on Rakuten Ichiba data, especially the proposed Modality-Attention merger method, (ii) compare the proposed model to recent multimodal/unimodal baselines on public benchmark datasets.   
\subsection{Datasets}
\label{sec:datasets}
We use in total five datasets. Two are large Rakuten Ichiba datasets for color and material attributes prediction. One is a Rakuten Ichiba data made available as part of a ENS data Challenge. Two are public benchmark datasets, namely MM-IMDB, and UPMC FOOD-101.  
%Three are publicly available datasets, including Rakuten Ichiba ENS data Challenge, MM-IMDB, and UPMC FOOD-101.
The basic statistics of these datasets are summarized in Table \ref{tab:datasets}.

%\subsubsection{Rakuten Ichiba datasets}
%In this section, we describe the datasets on which we conducted our experiments (see Table~\ref{tab:datasets}). The Rakuten Ichiba Color and Material datasets are private, the other 3 are publicly available. 
% We run experiments on these, following the benchmarks introduced in \cite{kiela2018efficient} and \cite{kiela2019}.
%We first describe the proprietary datasets made available to us. 
\begin{table}
\caption{Multimodal dataset statistics}
\label{tab:datasets}
\begin{tabular}{@{}lllc@{}}
\toprule
Name & \#Samples & Task type & \#Classes \\
\midrule
Rakuten Ichiba Material           & 6 725 281 & Multi-label & 57 \\
Rakuten Ichiba Color              & 6 460 720 & Multi-label & 19 \\
Rakuten ENS Color                 & 250 000   & Multi-label & 19 \\
\hline
MM-IMDB \cite{arevalo2017gated}   & 25 959    & Multi-label & 23 \\
UPMC FOOD-101 \cite{wang2015food} & 101 000   & Multi-class & 101 \\
\bottomrule
\end{tabular}
\end{table}

\paragraph{\textbf{Rakuten Ichiba Color and Material}}

%All items in the Rakuten Ichiba marketplace belong to a catalog, which consists in a taxonomy tree. Each node in the tree denotes a particular category.
Our data is part of the Rakuten Ichiba catalog and belongs to the following five categories: \emph{Men's Fashion}, \emph{Sports \& Outdoors}, \emph{Underwear}, \emph{Socks \& Sleepwear} and \emph{Women's Fashion}, which form a ``Fashion'' subset of the catalog. In order to create the \textit{Color} and \textit{Material} attribute datasets, we draw products belonging to this Fashion subset.
Every product in the database has one or more images, a title, a description and a list of attribute values referenced by external merchants that serve as ground truth. 

\paragraph{Labels} The labels are the \textit{attribute values}, which are provided by some merchants. For many products however the attribute values are missing. 
To fit and evaluate our model, we consider only products with at least one merchant attribute value. Every product can exhibit multiple attributes simultaneously. The Color dataset counts 19 attribute values, while the Material dataset has 57.

\paragraph{Image}
A product can have multiple images, which are ordered. The first image often contains sufficient information about the product. The remaining images usually depict either different views of the same product or embedded text giving details about the product. For merchants, it is mandatory to provide at least one image to add an product to the marketplace. We consider the first image only. %To bypass this rule, some merchants put placeholder images (images with "no image" written). Around 2\% of the items have such placeholder images, and these provide no relevant signal.

\paragraph{Text}
Products come with titles and descriptions, which we combine to form our text modality for every sample. Titles are shorter and used for display above a product, and are typically one sentence long, while descriptions contains more text and specifications of the product.   %We combine these two sources of information as described in section \ref{sec:text_only} before feeding them into our BERT encoder.

\paragraph{Dataset splits} %The Color dataset contains 6 460 720 samples and the Material dataset contains 6 725 281 samples.
Each dataset is divided into train and test splits, using the iterative stratification for multi-label data approach of \cite{sechidis2011stratification} and \cite{pmlr-v74-szymanski17a}. The number of samples in the train and test splits for Color are: 6 395 456 and 65 264, and for Material are: 6 537 277 and 188 004.\\

\paragraph{\textbf{Rakuten Ichiba - ENS Data Challenge}}~
A publicly available color prediction dataset from the Rakuten Ichiba marketplace was published for the 2021 \'Ecole Normale Sup\'erieure data challenge\footnote{\url{https://challengedata.ens.fr/participants/challenges/year/2021}}. 
%It contains 250k items and covers the same 19 colors as the whole dataset.
Unlike the larger Rakuten datasets, this dataset comprises images from all categories of the Rakuten Ichiba catalog.

%\subsubsection{Public benchmarking datasets}

\paragraph{\textbf{MM-IMDB} \cite{arevalo2017gated}}
% {\footnote{\url{http://lisi1.unal.edu.co/mmimdb/}}
This dataset includes about 26k movies. We use the plot and poster information. This dataset consist of 15552 train, 2608 dev and 7799 test samples. 
%The movies are categorized into 27 genres. However, 4 of these genres are only present in the test set or have less than 5 samples.
Following previous work \cite{singh2020, kiela2019}, we retain the 23 genres that are present across all data splits.

\paragraph{\textbf{UPMC FOOD-101}\cite{wang2015food}}
% \footnote{\url{http://visiir.lip6.fr/\#demo}} 
This dataset has 101k food images and associated text recipes, which are equally distributed in 101 classes. The recipes are in markdown format. 
%The recipes are crawled from different websites, and the HTML source code is cleaned by removing HTML tags resulting in the markdown training data.
Similar to previous work \cite{kiela2019}, we discard examples for which the text information is missing. We use the data splits made available by the authors of the dataset. 

%\paragraph{\textbf{MAE}}\cite{Logan2017MAE}
%This public e-commerce dataset contains 2.2 million products crawled from various marketplaces, each product having at least one image for a total of 4.0 million images. Product are represented by attribute-value pairs, which total to 7.6 million pairs for 2.1k unique attributes and 23.6k values. As preprocessing, we...
%# products 2.2 m
%# images 4.0 m
%# attribute-value pairs 7.6 m
%# unique attributes 2.1 k
%# unique values 23.6 k

%The original authors have noted that some recipes do not hold information relevant to describe the dish depicted by the image. These might be considered as noisy samples on which a multi-modal model would have to rely on the image.

% Details of results on Ichiba data, and possibly on similar publicly available datasets
% Notes on optimization, pretraining, hyperparameters used, etc

%{\color{blue}TODO}

%- Go over the chosen best architectures
%- Consider simplicity/off-the-shelf-y of our models compared to e.g. data processing in multimodal bitransformers, or pretraining in visio-linguistic pretraining

% The results we obtained are shown in Table~\ref{tab:upmc-food-101}, \ref{tab:mm-imdb}. 
%\subsection{Competing Methods} TODO afer Yako-san submission. Describe briefly the different competing methods.

\subsection{Evaluation Metrics}
\label{metrics}
During our model development we use common metrics for multi-label classification, namely F1-score, and aggregated F1-score per-category or per-label. However, for the purpose of pushing our model into production, it turns out that these metrics do not align well with the business needs. 
% These are also the metrics we use for the two public datasets on which we are testing our methods.
%However, the metric to certify putting our model in production is different, as this metric needs to be aligned with business needs.
%It is usually not mentioned in research papers, but common data science metrics used for research purposes are often not meaningful from a business perspective. A metric like F1-score can even miss what is most important for a business.
In our case the goal is to improve \emph{attribute coverage} -- the percentage of products that are assigned attributes -- while maintaining a high precision (of $95\%$). This is important since wrong predictions can render products unreachable.
%, and also lead to merchant complaints if their products have incorrect attribute values. 
To this end we adopt as metric \emph{Recall at $95\%$ precision} ($R@P95$), which is a good proxy for our objective.  

To assess performance on the public benchmark datasets we use the same metrics as the works we are comparing to. While for FOOD-101 and MM-IMDB the task is not attribute extraction, the means to achieve this task is multimodal classification, and these two datasets are commonly used as such. For FOOD-101 we use classification accuracy, which is a good measure given the task -- Multi-Class -- and the fact that the classes have the same number of samples. For MM-IMDB which is a Multi-Label dataset we use Macro-F1, Micro-F1, and Weighted-F1.

\subsection{Experimental Settings}
For Rakuten ichiba data, we use BERT pre-trained on Japanese Wikipedia \cite{cltohoku2021bert}. The model architecture is identical to original BERT base model, with $768$ dimensions of hidden states, $12$ layers, and $12$ attention heads. Tokenization takes into account latin characters as well as the three scripts in Japanese language: kanji, hiragana, and katakana. For the other datasets, with English texts, we use pre-trained BERT base model (uncased). We use the implementations available on HuggingFace \cite{wolf-etal-2020-transformers}, where more details regarding the pre-training, tokenization, and model architecture are available\footnote{\url{https://huggingface.co/cl-tohoku/bert-base-japanese-whole-word-masking}\\\url{https://huggingface.co/bert-base-uncased}}. As input to the classifier, we take the sum of the tokens' representations. We find that it gives comparable or superior performance compared to using the embedding of the [CLS] token.  

For images, we use DenseNet-121, with $121$ layers, pre-trained on the ImageNet (ILSVRC2012) dataset. The dimension of the output image representation is $D = 2048$. %When there are multiple images associated with a product, we encode the first one, which is usually the most representative of the item.

We use a held-out validation set to choose the values of different hyperparameters. For all the experiments we run, we set the batch size to $64$. In a pilot study we found that using other batch sizes (e.g., $32$, $128$) gives comparable results. We train all our models using the ADAM optimizer. For the text and multimodal models, we use a learning rate scheduler combining linear warmup with Cosine annealing. The maximum learning rate value is $2e^{-5}$ for Rakuten Ichiba data and $5e^{-5}$ for the remaining datasets. For the image-only model, we set a constant learning rate to $1e^{-4}$. For the number of epochs we use the following search space $\{5,\ldots,20\}$ with steps of $5$. For the modality-attention regularization parameter $\lambda$, the search space is  $\{0.0, 1e{-4},5e{-4},\ldots,1e{-1}, 5e{-1}\}$. We find on our datasets that beyond a value of $5e{-1}$ for $\lambda$, the attention distributions much almost perfectly the uniform distribution.

\subsection{Results on Rakuten Datasets}
\label{sec:main-results}

The results of our investigations on the Rakuten datasets are summarized in this section. The main findings are as follows.

\begin{figure*}[!t]
    \centering
    \includegraphics[scale=0.36]{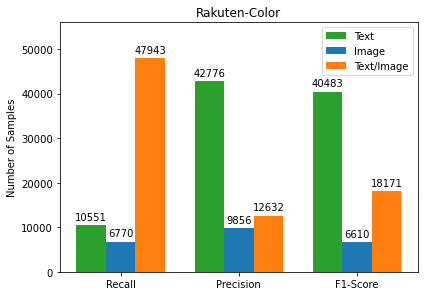}
    \quad
    \includegraphics[scale=0.37]{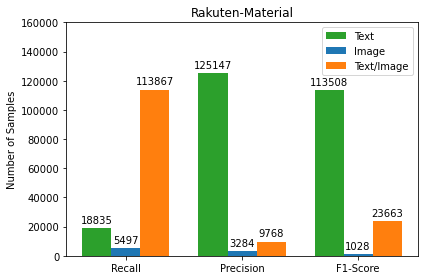}
    \quad
    \includegraphics[scale=0.37]{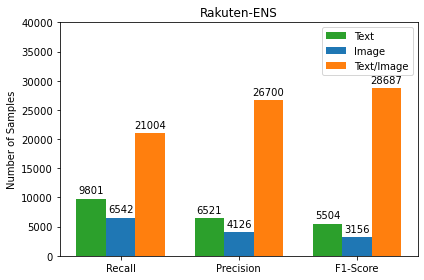}
  \caption{\small The number of samples (products) on which each modality performs the best. The Text/Image legend means both modalities perform equally. Clearly no modality performs the best on all samples.}
  \label{fig:best_modality}
\end{figure*}

\emph{Cross-Modality comparisons.} Table \ref{tab:modality-comparison} shows that, on average, the text-only model offers substantially higher performance than the image-only model. The gap is more important on the material dataset. This can be explained in part by the characteristics of the images we deal with, e.g., containing multiple instances of the same product under different attribute values (see Figure \ref{fig:max-system} for instance) and overlaid with text, which can confuse the image model. Another explanation is the challenging nature of our tasks. For instance, identifying material type from a product image may be difficult even for humans. For per-sample performance, Figure \ref{fig:best_modality} shows that there are many cases in which the image model does better than the text model. Hence, both modalities matter and they can complement each other. This is empirically supported by the results of our multimodal model, third row for each dataset in Table \ref{tab:modality-comparison}, which offers the best performance in all cases.

\emph{Importance of the proposed merging method.} The proposed modality-attention merger outperforms the vanilla concatenation method as shown in Table \ref{tab:merger-arch}. Moreover, simply concatenating the text and image modalities may results in sub-optimal performance. For instance, referring to tables \ref{tab:modality-comparison} and \ref{tab:merger-arch}, on Rakuten color, the multimodal method with the Concat merger stands behind the text only model. This provides further support for the importance of our merging strategy, which offer more flexibly for combining modalities. Interestingly, from Figure \ref{fig:best_modality} and Table \ref{tab:merger-arch}, it seems that the modality attention merger offers the most important improvements when there is more difference between the two modalities -- lower orange/third bar for Precision and F1-Score.        

\begin{figure*}[!t]
    \centering
    \includegraphics[scale=0.39]{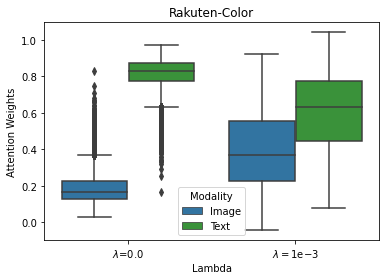}
    \quad
    \includegraphics[scale=0.39]{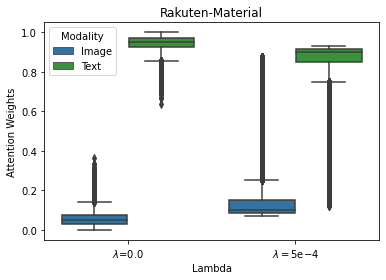}
    \quad
    \includegraphics[scale=0.39]{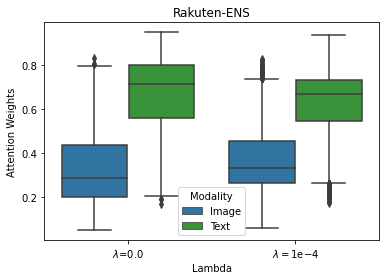}
  \caption{\small Attention weights distribution across modalities on the Rakuten datasets. On average the model assigns higher weights to the text modality than the image. The KL-regularization alleviates modality collapse on Rakuten-Material.}
  \label{fig:attention_weights}
\end{figure*}

\emph{Modality Collapse and Importance of the KL-regularization.}
We observe modality collapse on the Rakuten-Material datatset. The model assigns very low weights for the image modality for almost all samples when the regularization parameter $\lambda=0.0$ (see Figure \ref{fig:attention_weights}). We suspect that this is caused by the important gap between the Text and Image modalities. That is, as shown in figure \ref{tab:modality-comparison}, Text is more relevant than image for material prediction in most cases. Hence, the model can be easily biased to rely on text only. This is analogous to class-size imbalance problem in classification. The proposed KL-regularization seems to alleviate posterior collapse on Material, which also translates to better performance in F1-Score and R@P95 as depicted in Table \ref{tab:merger-arch}.

%Multi-modal models inherently require more compute, both during training and inference. We train different models on our datasets in order to quantify the contribution of each modality. The results are shown in Table~\ref{tab:modality-comparison}. For all datasets, we observe an increase in model performance when using both text and image modalities. 

%Looking at the performance of the individual modalities, we see that using just the image modality leads to a worse performance compared to training on just the text modality. This can be explained by the fact that words describe concepts unambiguously and with a greater starting abstraction level than pixels in an image. Further, combining the two modalities leads to an increase in performance as compared to a text only model. This demonstrates that the image-only model provides complementary information to text-only model, leading to a net increase in performance.

%
\begin{table}[!t]
\caption{Modality comparison. The last row for each dataset corresponds to the proposed multimodal model with the modality-attention merger and the KL-regularization.}
\label{tab:modality-comparison}
\begin{tabular}{@{}lccccc@{}}
\toprule
Dataset & Image & Text & F1-Score (\%) & R@P95 (\%)\\
\midrule
       & \ding{51} & \ding{53} & 61.8 & 8.0 \\
Rakuten-Color & \ding{53} & \ding{51} & 79.3 & 57.3  \\
       & \ding{51} & \ding{51} & \textbf{84.5} & \textbf{66.0}  \\
\midrule
         & \ding{51} & \ding{53} & 55.5 & 5.0  \\
Rakuten-Material & \ding{53} & \ding{51} & 84.6 & 60.4  \\
         & \ding{51} & \ding{51} & \textbf{87.0} & \textbf{68.0}  \\
\midrule
            & \ding{51} & \ding{53} & 61.5 & 13.5 \\
Rakuten-ENS & \ding{53} & \ding{51} & 69.2 & 35.0 \\
            & \ding{51} & \ding{51} & \textbf{76.6}& \textbf{43.2} \\
\bottomrule
\end{tabular}
\end{table}
{
\setlength{\tabcolsep}{3pt}
\begin{table}[!t]
\caption{Merging methods comparison.} \label{tab:merger-arch}
\begin{tabular}{@{}p{0.15\linewidth}p{0.46\linewidth}cc@{}}
% \begin{tabular}{@{}llcc@{}}
\toprule
Dataset & Merger & F1-Score (\%) & R@P95 (\%) \\
\midrule
%Rakuten        & Concat & 78.58 & 52.55 \\
%Color          & Modality-Attention & 84.25 & 65.49 \\
%               & Reg. Modality-Attention & \textbf{84.54} & \textbf{65.96} \\ 
Rakuten        & Concat & 78.6 & 52.6 \\
Color          & Modality-Attention ($\lambda=0.0$) & 84.3 & 65.5 \\
               & Modality-Attention ($\lambda=1e-3$) & \textbf{84.5} & \textbf{66.0} \\ 
\midrule
%Rakuten           & Concat  & 85.43 & 62.51 \\
%Material          & Modality-Attention & 84.87 & 61.11 \\
%                  & Reg. Modality-Attention & \textbf{87.00} & \textbf{68.01} \\
Rakuten           & Concat  & 85.4 & 62.5 \\
Material          & Modality-Attention ($\lambda=0.0$) & 84.9 & 61.1 \\
                  & Modality-Attention ($\lambda=5e-4$) & \textbf{87.0} & \textbf{68.0} \\
                  
\midrule
%Rakuten         & Concat & 75.80 $\pm$ 0.16 & 43.16 \\
%ENS             & Modality-Attention & 76.22 $\pm$ 0.07 & \textbf{44.52} \\
%                & Reg. Modality-Attention & \textbf{76.66 $\pm$ 0.16} & 43.24 \\
                
Rakuten         & Concat & 75.8 & 43.2 \\
ENS             & Modality-Attention ($\lambda=0.0$) & 76.2 & \textbf{44.5} \\
                & Modality-Attention ($\lambda=1e-4$) & \textbf{76.7} & 43.2 \\
                %& Rakuten-Color pretraining & 77.80 $\pm$ 0.15 \\
\bottomrule
\end{tabular}
\end{table}
}
%

%\paragraph{Impact of the proposed modality merging approach}

%To assess the impact of the proposed modality-attention merging method (with and without regularization), we compare it to the vanilla concatenation approach. The results are depicted in Table~\ref{tab:merger-arch} and \ref{tab:merger-arch-public}. Our modality merging methods show comparable or superior performance to simple concatenation, thereby demonstrating the benefits of our approach. We also observe that in several cases, the proposed regularization strategy allows the modality-attention merging to reach even higher results. Recall that the purpose of this regularization is to prevent degenerate solutions, in which the model forgets one modality, i.e., always assigns a zero weight to one modality.  

\emph{Qualitative results.}
Figure \ref{fig:attention_weights} depicts the distribution of the attention weights learned by our model on the Rakuten datasets. Interestingly, we observe that, on average the model assigns higher weights to the text modality than the image. This is inline with our results showing that the text-only models outperform the image-only models. Nevertheless, the model also attends the image especially on Rakuten ENS and Color datasets. The boxplots also show that there are several cases where the model assigns a very high weight to one modality. Figure \ref{fig:qualitative_example_attention} shows two qualitative examples of such extreme cases. For instance, in the second example the model seems to rely on the image because the color information is missing in the text. Hence, in addition to improving performance, the proposed modality merging approach allows to learn a more interpretable model. 

\emph{Human evaluation}.
If $R@P95$ drives model selection forward in offline experiments, our multimodal model with the modality attention merger has also gone through a human evaluation before production. To this end, a dataset of 5000 randomly sampled products is made by the relevant business unit. Sampling is carried out using iterative stratified multilabel sampling, in order to have a balanced distribution of all existing attributes values for the considered categories. These samples are then fed into our model for attribute prediction.
%Second, the selected model was run against those samples.
Finally, the product image, text, and our model predictions are shown to a human annotator who, is asked to answer ``yes'' or ``no'' to the following question: \emph{For that product, are the attributes predicted by the model correct?} 
%Third, a human annotator was shown the product image and text as well as the model prediction, and was asked the yes/no question "For that product, are the attributes predicted by the model correct?".
Our model achieves an accuracy of $98\%$ based on this human evaluation, and thereby validating its deployment in Rakuten Ichiba. 
%An accuracy was computed according to this binary output, and for our final model, it reached a level of $98\%$, validating its productionization.
%
%

%

%\begin{figure}[!t]
%  \centering
%  \includegraphics[scale=0.5]{figures/attention_boxplots_material.png}
%  \caption{\small Attention weights distribution across modalities and different values of $\lambda$.}
%  \label{fig:attention_weights_material}
%\end{figure}

%
%\begin{figure}[!t]
%  \centering
%  \includegraphics[scale=0.37]{figures/attention_boxplot.png}
%  \caption{\small Attention weights distribution across modalities on the Rakuten ENS dataset. On average the model assigns higher weights to the text modality than the image. However, no modality is discarded by the model highlighting the importance of both.}
%  \label{fig:attention_weights}
%\end{figure}

%
%{\setlength\tabcolsep{2pt}
%\begin{table}[h]
%\caption{Modality comparison on public benchmark datasets}
%\label{tab:modality-comparison-public}
%\begin{tabular}{@{}lccccc@{}}
%\toprule
%Dataset & Image & Text & F1-Score (\%) & Accuracy\\
%\midrule
%          & \ding{51} & \ding{53} &  & 68.8\\
%UPMC Food-101  & \ding{53} & \ding{51} &  &  87.6\\
%          & \ding{51} & \ding{51} &  &  \textbf{93.6}\\
%\midrule
%          & \ding{51} & \ding{53} & 40.7 &   \\
%MM-IMDB  & \ding{53} & \ding{51} & 66.9 &  \\
%          & \ding{51} & \ding{51} & \textbf{68.4} &  \\
%\bottomrule
%\end{tabular}
%\end{table}
%}

\subsection{Results on Public Benchmarks}
We now compare our multimodal model with several baselines on public benchmarks. We repeat every experiment five times with different seeds and report the average and standard deviation results for each model. We reuse results shared in other works in tables ~\ref{tab:mm-imdb} and \ref{tab:upmc-food-101}. The baselines are briefly described at the end of the section. 

{
\setlength{\tabcolsep}{3pt}
\begin{table}
\small
\caption{MM-IMDB Results}
\label{tab:mm-imdb}
% \resizebox{9cm}{!}{
\begin{tabular}{@{}lp{0.3\linewidth}cccc@{}}
 \toprule
 Type & Model & Macro-F1 & Micro-F1 & Weighted-F1 \\
 \midrule
 \emph{Single Modality} & DenseNet-121 & 20.3 & 45.9 & 40.7\\
  %\midrule
  & BERT & 58.4 & 68.1 & 66.9\\
 \midrule
   & MMBT \cite{kiela2019} & \textbf{61.6$\pm$ 0.2}& 66.8 $\pm$0.1 & \\
  &  CentralNet \cite{Vielzeuf2018}  & 56.1 & 63.9 & 63.1   \\
  \emph{Early Fusion}   & ELSMC \cite{kiela2018efficient} & & & 62.3$\pm$ 0.2 \\
    %MFAS \cite{perezrua2019} & 55.68 & & 62.50  \\
   & MFAS \cite{perezrua2019} & 55.7 & & 62.5  \\
    %BM-NAS \cite{yin2021} & & & 62.92  \\
  &  BM-NAS \cite{yin2021} & & & 62.9  \\

\midrule
& GMU \cite{arevalo2017gated} & 54.1 & 63.0 & 61.7 \\
% Visio-linguistic pretraining \cite{singh2020} & 60.02 & 68.14 &  \\
 &VLP \cite{singh2020} & 60.0 & 68.1 &  \\
% Multimodal Sets \cite{reiter2020} & 61.33 & 67.73 &  \\
\emph{Late Fusion} &Multimodal Sets \cite{reiter2020} & 61.3 & 67.7 &  \\
& PM+MO  \cite{armitage2020} & 54.9 & 62.0 & 61.7  \\
 %Modality specific distillation \cite{jin2021} & 53.12 & 63.00 &  \\
& MSD \cite{jin2021} & 53.1 & 63.0 &  \\
 \midrule
  & \textbf{Concat} & 61.2 $\pm$ 1.2 & \textbf{69.4 $\pm$ 0.5} & \textbf{68.4} $\pm$ \textbf{0.7}\\
 Ours & Mod.-Att. ($\lambda = 0$) & 60.2 $\pm$ 1.2 & 69.0 $\pm$ 0.9 & 67.7 $\pm$ 1.1 \\
  & Mod.-Att. ($\lambda = 5e-4$) & 60.6 $\pm$ 0.7 & 69.2 $\pm$ 0.4 & 68.1 $\pm$ 0.5 \\
 \bottomrule
\end{tabular}
% }
\end{table}
}

{
\setlength{\tabcolsep}{3pt}
\begin{table}
\small
\caption{UPMC FOOD-101 Results}
\label{tab:upmc-food-101}
\begin{tabular}{@{}cp{0.6\linewidth}ccc@{}}
 \toprule
Type & Model & Accuracy \\
\midrule
\emph{Single Modality} & DenseNet-121 & 68.8\\
%\midrule
  & BERT & 87.6\\
 \midrule
 \emph{Early Fusion} &MMBT \cite{kiela2019} & 92.1 $\pm$ 0.1\\
 & ELSMC \cite{kiela2018efficient} & 90.8 \\
 \midrule
\emph{Late Fusion} & Score fusion \cite{wang2015food} & 85.1 \\
%  JDNet \cite{zhao2020jdnet}  & 91.2 \\
 %DMSRC \cite{abavisani2020}  & 92.75 \\
 & DMSRC \cite{abavisani2020}  & 92.8 \\
 \midrule
% Concat(ours) & 93.6 $\pm$ 0.1 \\
% ConcatAttention & 93.5 $\pm$ 0.06\\
%\textbf{Reg. Modality-Attention} & \textbf{93.7 $\pm$ 0.03} \\
     & Concat & 93.6 $\pm$ 0.1 \\
Ours & Modality-Attention ($\lambda = 0.0$) & 93.5 $\pm$ 0.06 \\
     & \textbf{Modality-Attention ($\lambda = 0.1$)} & \textbf{93.7 $\pm$ 0.03} \\
\bottomrule
\end{tabular}
\end{table}
}
On MM-IMDB (Table~\ref{tab:mm-imdb}), our multimodal architecture with regardless of the merging method offers substantially higher performance than the other competing methods in terms of Micro-F1 and Weighted-F1. In macro-F1, our model is slightly below the MMBT and Multimodal Sets methods. The performance is however tight, and the differences do not seem to be significant. We observe a similar trend on UPMC FOOD-101 in Table \ref{tab:upmc-food-101}. These results are appealing as our model is relatively simpler than most of the multimodal baselines we compare with. 

Note that on these two benchmark datasets, we do not observe important differences between the modality-attention and concat mergers. We attribute this behavior to the relatively less challenging nature of these datasets. In fact, compared to our Rakuten data, on MM-IMDB and Food-101 there is less discrepancy/unbalance between the performance (or importance) of the two modalities.       

%Table \ref{tab:upmc-food-101} show that on UPMC FOOD-101, the proposed multimodal architecture under the different mergers offers the best classification accuracy among baselines, which provides positive supports for the importance of our approach.

\paragraph{Brief description of baselines.}
%We go over some technical specifications of the methods previously used in the literature on these tasks and this data. 
%We separated early fusion and late fusion methods as explained in the related work section.
CentralNet \cite{Vielzeuf2018} uses a representation of text that comes from word2vec, and an image representation coming from a VGG-16 pretrained on ImageNet. ELSMC, Efficient large-scale multimodal classification\cite{kiela2018efficient} uses ResNet and FastText, and reach lower performances than our model on both datasets. MFAS \cite{perezrua2019} and next BM-NAS \cite{yin2021} has end-to-end neural architecture search allowing to build representations that draw from intra-modal and inter-modal interactions, and one of these is equivalent to multi-head attention. VLP, Visio-linguistic pretraining \cite{singh2020} uses two multimodal transformers-based architectures, VisualBERT and VilBERT, and vary the pretraining beforehand. Multimodal Sets \cite{reiter2020} uses sets instead of vectors as inputs, and add a bias to the classification layer to help with the imbalance. PM+MO \cite{armitage2020} regularize the multimodal objective with variational inference. They combine modalities encoded using VGG16 and word2vec. MSD, Modality-Specific Distillation \cite{jin2021}, uses VisualBERT to combine the two modalities and perform knowledge distillation on both independently. Score Fusion \cite{wang2015food} is the baseline on FOOD-101. It uses VGG-19 for images and Tf-Idf embeddings for text representation, and combines the output scores of two classifiers. MMBT \cite{kiela2019} combines textual and visual tokens into a single transformer. This method slightly better than ours on MM-IMDB, but reaches lower results than ours on FOOD-101. DMSRC \cite{abavisani2020} builds sparse representations using autoencoders specifically trained for this task end-to-end. For the image modality, they use conventional convolution/deconvolution stacks. This tuned architecture gets second-best results on FOOD-101.

\section{Conclusion}
In this work, we consider product attribution extraction from text and images. We find that achieving good performance on this task is not direct, but rather the results of several refinements and systematic investigations into several practical research questions revolving around multimodality, such as \emph{which model to rely on}, \emph{which modality performs best} and \emph{how we should combine multiple modalities}. Cross-modality comparisons on Rakuten data, reveal that the best preforming modality may be sample-dependent, which inspired us to develop a flexible merging method allowing the model to choose the modality to rely on the most for every product. We further propose a principled regularization scheme to mitigate modality collapse, which seems to occur when there is a very important gap between the performance of the text and the image modalities. Moreover, we also discuss and characterize the situations in which the proposed modality-attention merger offers the most significant improvements. Our multimodal model with the proposed merging method has successfully passed human evaluations, and it is currently deployed in Rakuten Ichiba. We hope the investigations presented in this paper will benefit and inspire future academic/applied research on the topics of multimodality and attribute extraction for e-commerce product.
%In this work we introduce a new and simple to implement approach for product attribute extraction from texts and images. We tackle attribute extraction as a supervised classification problem. Although classification tasks are well studied and there is a wealth of literature on this subject, leveraging existing models for real-world use cases uncovers significant hurdles. We focus on building efficient single-modality models, as well as on how to effectively combine different modalities. We use a modality-attention merging approach and note that combination of modalities often yields \emph{modality collapse}. Hence, we introduce entropy regularization to mitigate this collapse. Our experimental results provide empirical evidence of the benefits of our contribution. The proposed modality-attention merging approach with entropy regularization achieves satisfactory results on Rakuten Ichiba data. Moreover, human evaluations show that our multimodal model is $98\%$ accurate, which allowed it to be deployed in the Rakuten Ichiba marketplace.   We also included several baselines on public benchmarks for multimodal classification, and shown that our model offered competitive performance. 

%\nocite{*}
%\begin{acks}
%We thank our colleagues at RIT Paris and RIT Boston. 
%\end{acks}

\bibliographystyle{ACM-Reference-Format}

\bibliography{bibliography}
\clearpage
\appendix

%\section{Effect of title and descriptions}
%\begin{table}[h]
%\caption{Item text experiments}
%\label{tab:ichiba-text}
%\begin{tabular}{@{}lcccc@{}}
%\toprule
%Dataset & Title & Description & F1-Score (\%) & R@P95 (\%) %\\
%\midrule
%Rakuten-Color     & \ding{51} & \ding{53} & 73.6 & 46.7 \\
%         & \ding{51} & \ding{51} & \textbf{79.3} & \textbf{57.3} \\
%\midrule
%Rakuten-Material  & \ding{51} & \ding{53} & 63.7 & 50.5 \\
%         & \ding{51} & \ding{51} & \textbf{67.5} & \textbf{60.4} \\
%\bottomrule
%\end{tabular}
%\end{table}

%Note that for efficiency purposes, we have considered training our text model from titles only. However, as can be seen from Table~\ref{tab:ichiba-text}, learning from both titles and descriptions improves performance substantially. 

%On the Rakuten Ichiba datasets, we have two sources of text information: title and description. The item title is generally shorter than the item description, and contains a summary of the item. The item description provides details and characteristics of the product. Intuitively we expect that using descriptions to train the model will bring about improvements in the model performance, as it will provide detailed information absent from item titles. The results for training with and without item descriptions are shown in Table~\ref{tab:ichiba-text}. The results show that using descriptions improves model performance for both of our tasks. However, using descriptions leads to longer input sequences. In the case of BERT, this is a quadratic increase in compute.

\end{document}